\documentclass[11pt]{article}
\usepackage{algorithmic}

\usepackage[]{EACL2023}

\usepackage{times}
\usepackage{latexsym}

\usepackage[T1]{fontenc}

\usepackage[utf8]{inputenc}

\usepackage{microtype}

\usepackage{inconsolata}

\usepackage{caption}
\usepackage{amsfonts}
\usepackage{amsmath}
\usepackage{graphicx}
\usepackage{subfigure} 
\usepackage{multirow}
\usepackage{booktabs}
\usepackage{pifont}
\usepackage{makecell}
\usepackage{xcolor}
\usepackage{CJKutf8}

\title{\emph{Original} or \emph{Translated}? On the Use of Parallel Data\\ for Translation Quality Estimation}

\author{Baopu Qiu$^{1}$\thanks{~~Work was done when Baopu was interning at JD Explore Academy.}~, \ Liang Ding$^{2}$, \ Di Wu$^{3}$, \ Lin Shang$^{1}$, \ Yibing Zhan$^{2}$, \ Dacheng Tao$^{2}$\\
$^{1}$Nanjing University \ $^{2}$JD Explore Academy \ $^{3}$University of Amsterdam\\
\texttt{qiubaopu@smail.nju.edu.cn}, \texttt{dingliang1@jd.com}
}

\begin{document}
\maketitle
\begin{abstract}
Machine Translation Quality Estimation (QE) is the task of evaluating translation output in the absence of human-written references. Due to the scarcity of human-labeled QE data, previous works attempted to utilize the abundant unlabeled parallel corpora to produce additional training data with pseudo labels. In this paper, we demonstrate a significant gap between parallel data and real QE data: for QE data, it is strictly guaranteed that the source side is original texts and the target side is translated (namely translationese). However, for parallel data, it is indiscriminate and the translationese may occur on either source or target side.
We compare the impact of parallel data with different translation directions in QE data augmentation, and find that using the source-original part of parallel corpus consistently outperforms its target-original counterpart. Moreover, since the WMT corpus lacks direction information for each parallel sentence, we train a classifier to distinguish source- and target-original bitext, and carry out an analysis of their difference in both style and domain. Together, these findings suggest using source-original parallel data for QE data augmentation, which brings a relative improvement of  up to 4.0\% and 6.4\% compared to undifferentiated data on sentence- and word-level QE tasks respectively.
\end{abstract}

\section{Introduction}

\begin{figure}[ht]

\includegraphics[scale=0.41]{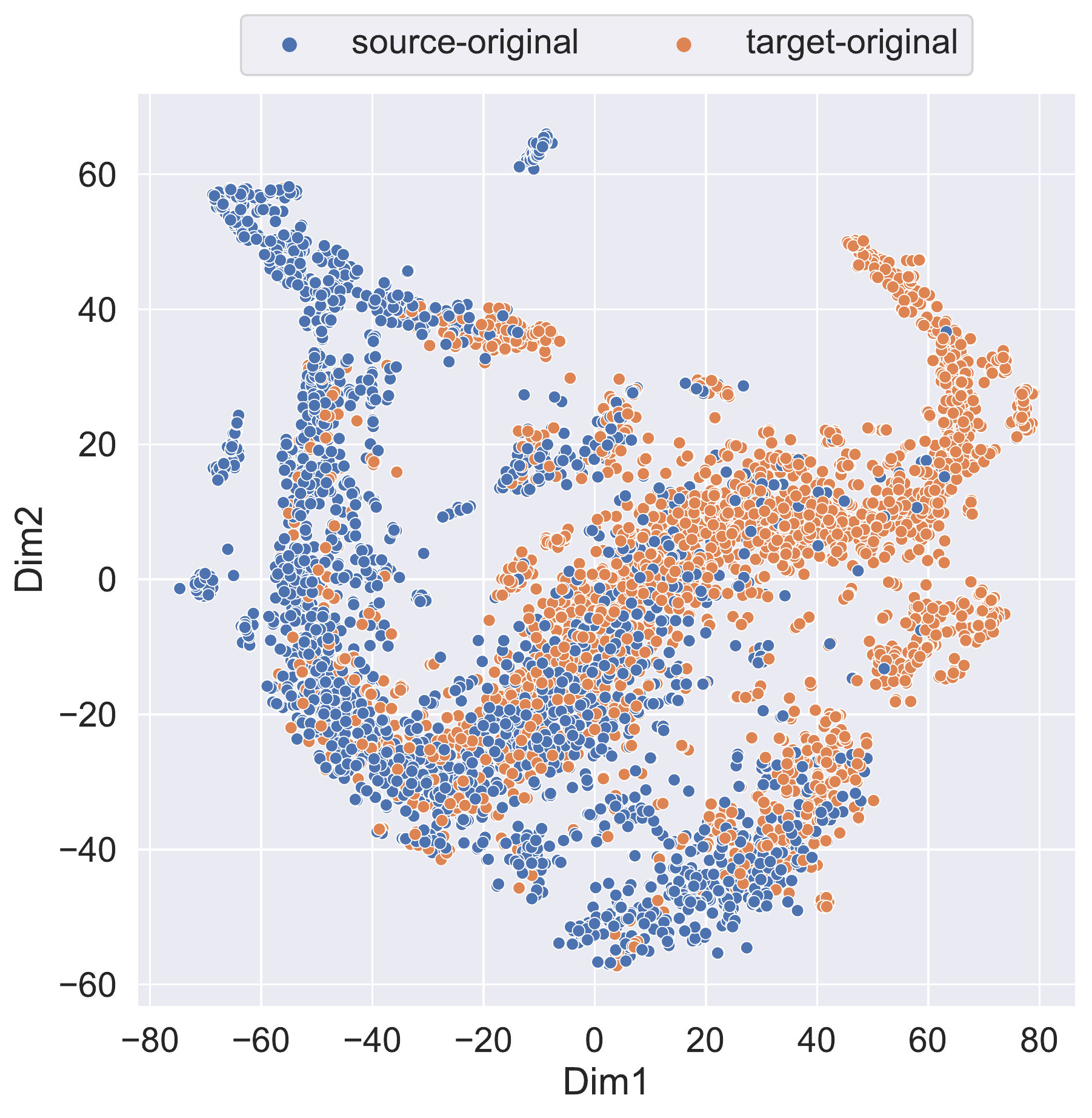}
\centering
\caption{\textbf{Visualization of target-side sentence embeddings in WMT22 En-Zh training dataset.} Source-/Target-original labels are produced by our translation direction classifier proposed in Section \ref{sec:classify}. We can observe the difference between the distribution of  ``source-original'' (target-side is \emph{translationese}) and ``target-original''(target-side is \emph{original text}).}
\label{fig:motivation}
\end{figure}

\begin{figure*}[ht]
\includegraphics[scale=0.63]{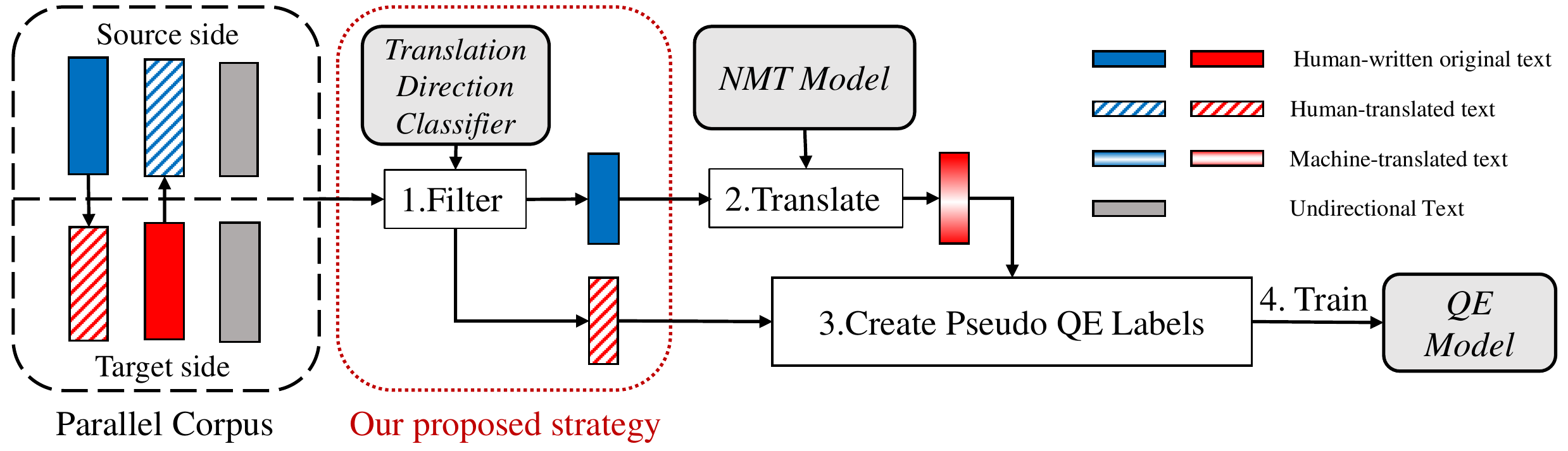}
\centering
\caption{\textbf{Pipeline of our proposed method.} We first train a translation direction classifier. Then we extract source-original bitext from the parallel corpus. The selected bitext is used to produce additional QE training data with pseudo labels. }
\label{fig:pipeline}
\end{figure*}

Machine Translation Quality Estimation (QE) focuses on evaluating the quality of machine-generated text without relying on pre-defined references. Due to the scarcity of supervised QE data, data augmentation methods have been playing an important role in QE. \citet{DBLP:journals/talip/KimJKLN17} proposed the predictor-estimator framework, in which a predictor is pretrained on a large amount of parallel data to provide bilingual features, and the estimator is finetuned on a small amount of gold label QE data. Recently in WMT QE shared task \citep{DBLP:conf/wmt/SpeciaBFFCGM20, specia2021findings}, while directly finetuning pretrained cross-lingual LMs (XLM, XLM-R) can already achieve satisfying results \citep{ranasinghe2020transquest,zerva2021unbabel}, most participants \citep{DBLP:conf/wmt/BaekKMKP20, DBLP:conf/wmt/WuWMWWWZYP20,zervafindings} still find it beneficial to use the WMT parallel corpus to produce additional training data with pseudo labels of the QE task, i.e. word/sentence-level quality prediction. However, there is a style and domain mismatch between pseudo QE data and real QE data: real QE data are labeled machine translations of the original text in the source language, while the pseudo QE data consist of translations of both \emph{original text} and \emph{translationese}\footnote{
Previous works \citep{DBLP:conf/wmt/ZhangT19,DBLP:conf/emnlp/GrahamHK20,riley2020translationese} refer to the term \emph{Translationese} as text translated to a target language either by human or by MT systems \citep{Gellerstam1986TranslationeseIS,gellerstam1996translations}. It is a counterpart to \emph{Original Text}, which refers to the originally human-written text in that language. }. As Figure~\ref{fig:motivation} shows, \emph{translationese} 
can be easily differentiated from the originally-written text.


In this paper, we attempt to alleviate the discrepancy between pseudo QE data and gold-label QE data by distinguishing source-original and target-original bitext in the parallel corpus.
We conduct an in-depth analysis on the effect of different translation directions of bitext  applied in generating pseudo labels for downstream QE tasks \emph{by}
\ding{182} Conducting preliminary experiments to reveal that data of original source and translationese target (i.e. \emph{source-original} bitext) is closer to real QE data in both style and domain in Section~\ref{sec:discrepancy}.
\ding{183} Training a classifier to distinguish different types of bitext in the WMT parallel corpus. Based on our pilot experiments in Section \ref{sec:classify}, content-dependent differences are more salient features to distinguish parallel sentence pairs originating from different languages compared to stylistic differences. 
\ding{184} Experimentally showing that selecting the source-original part of parallel corpus brings a relative improvement of  4.0\% and 6.4\% compared to undifferentiated data on sentence- and word-level QE tasks respectively.


\section{Motivation}

The overall framework of our proposed method is displayed in Figure ~\ref{fig:pipeline}. We focus on analyzing the effect of different types of parallel data to improve the standard QE data augmentation pipeline.

\subsection{Task definition}
Given an input source text $\texttt{src}=\{x_1, ..., x_n\}$ and machine-translated output $\texttt{sys}=\{y_1,..., y_m\}$, a QE model outputs:
$$\{s, Q\} = \textsf{QE}(\texttt{src}, \texttt{sys})$$
where $s$ is a sentence-level quality score, $Q$ is a set of word-level quality tags  in which $q_i\in\{\texttt{OK}, \texttt{BAD}\}$ are binary labels\footnote{In the WMT QE track, the task of predicting post-editing efforts includes predicting both sentence-level and word-level quality labels. Sentence-level scores measure the post-editing effort while word-level tags indicate whether the token in a specific position is translated correctly.}. 

\subsection{Data Augmentation in QE}
Based on statistics of all the submissions to the WMT19-21 QE shared task, we observed a prevalence of data augmentation methods that apply parallel data to produce extra QE training data (77.8\% for the post-editing effort subtask, 42.3\% for the direct assessment subtask, 77.8\% for other subtasks). 


The most common QE data augmentation method utilizes the target side of parallel data as the reference translation $\texttt{ref}$ \citep{DBLP:conf/wmt/BaekKMKP20, DBLP:conf/wmt/WuWMWWWZYP20,zervafindings}. An MT model is used to translate the source side to \texttt{sys}, then pseudo QE labels can be computed via reference-based MT metrics \citep{freitag-etal-2021-results}\footnote{Word-level tags are acquired when calculating HTER score \citep{snover2006study,specia2009estimating}}.
$$\{s', Q'\} = \textsf{Metric}(\texttt{sys}, \texttt{ref})$$ 

Intuitively, we want \textbf{the pseudo QE data to be as close to real QE data as possible}. Real QE data is naturally related to the source-original part of parallel data, since QE is the task evaluating translations of the original source text, while the target side is always translationese. Thus, we focus on the translation direction of the WMT parallel corpus, which is often utilized by QE-shared task participants to produce extra training data. 

\subsection{Translation Direction in Parallel Data} 
Researchers have observed that most parallel corpora are composed of translations from both directions \citep{riley2020translationese,DBLP:conf/acl/WangTTSSL21}, which can be either source-original (source-side is original text) or target-original (source-side is translationese). Translation direction has been proved to exert a subtle influence in MT performance \citep{caswell2019tagged,marie-etal-2020-tagged} and lexical choice~\cite{ding2021understanding,ding-etal-2021-rejuvenating}.

\begin{figure}[h]
\centering
\includegraphics[scale=0.54]{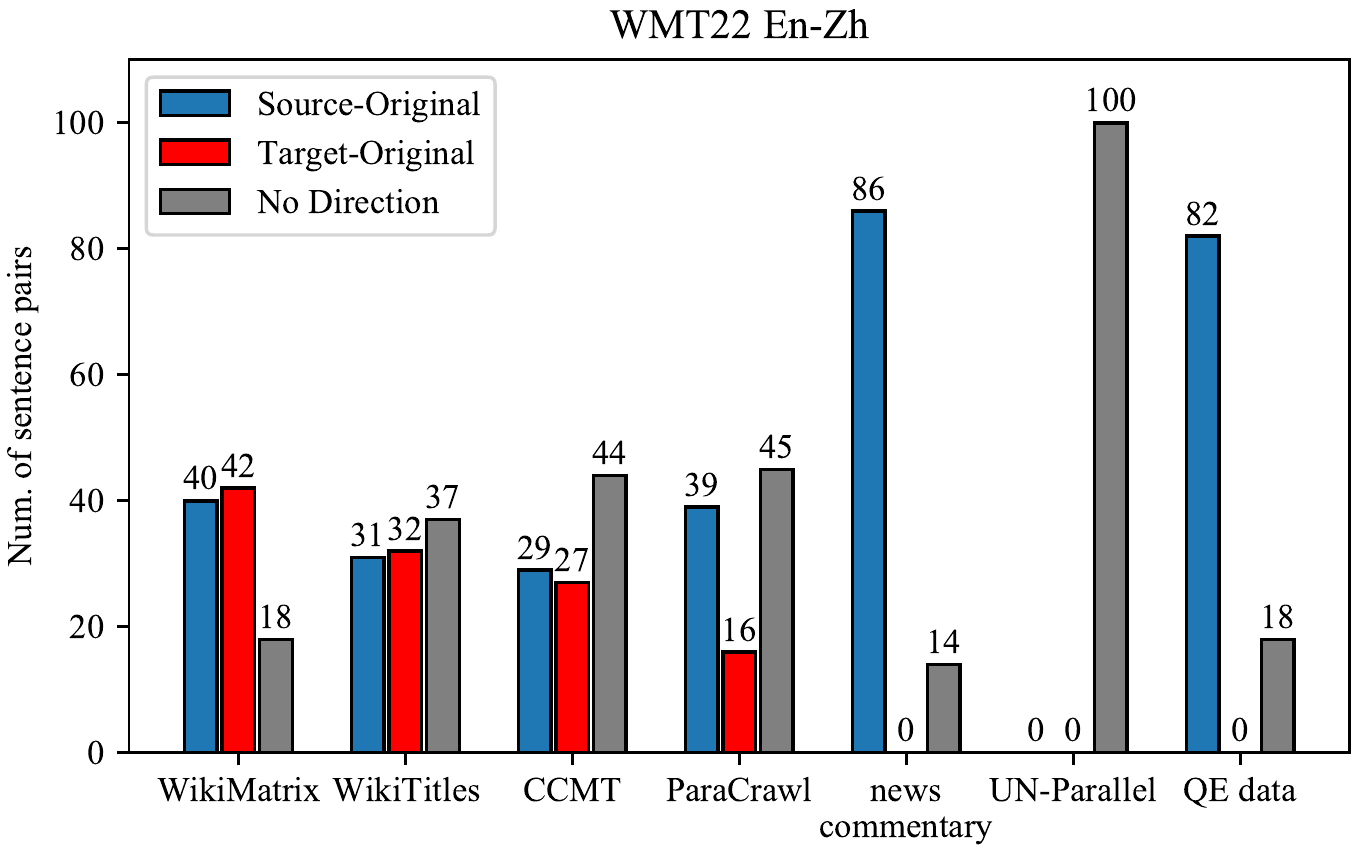}
\caption{\textbf{Translation direction statistics} of each parallel corpus in WMT22 En-Zh training set.}
\label{fig:para}
\end{figure}

To investigate translation direction in the parallel data widely-adopted for QE data augmentation, we randomly sampled 100 sentence pairs from each sub-corpus of the WMT22 En-Zh training dataset for the general MT task\footnote{https://statmt.org/wmt22/translation-task.html}. We annotate their translation direction manually, according to the human heuristics of the named entities and language style. For example, 
in En-Zh sentence pair ``Yellowstone National Park spans approximately 2.2 million acres$\rightarrow$\begin{CJK}{UTF8}{gkai}黄石国家公园占地约220万英亩\end{CJK}'', the entity ``Yellowstone National Park'' is obviously more probable to occur in English-original text, therefore, the direction will be marked as ``source original''.
For those without explicit translation direction, we label them as ``no direction''. As shown in Figure~\ref{fig:para}, most of the parallel corpus is composed of directional sentence pairs. Exceptions include the UN-Parallel corpus, in which all sentences are from aligned official documents of the United Nations. 

\subsection{Discrepancy between Source-original and Target-original Bitext}
\label{sec:discrepancy}
We conduct pilot experiments to compare real QE data with both parts of parallel data on their stylistic properties and vocabulary distribution. 

\paragraph{Stylistic Differences}
To investigate stylistic features, we compute the lexical variety, measured by token-type-ratio ($TTR=\frac{num.\, of\, types}{num.\, of\, tokens}$), as well as lexical density, measured by the ratio of content words. These two metrics have been proposed by \citet{DBLP:conf/mtsummit/Toral19} as stylistic features to quantify the degree of translationese in text. \emph{Lower} lexical density and variety indicate \emph{higher} degree of translationese.

\begin{figure}[t]
\centering
\includegraphics[scale=0.33]{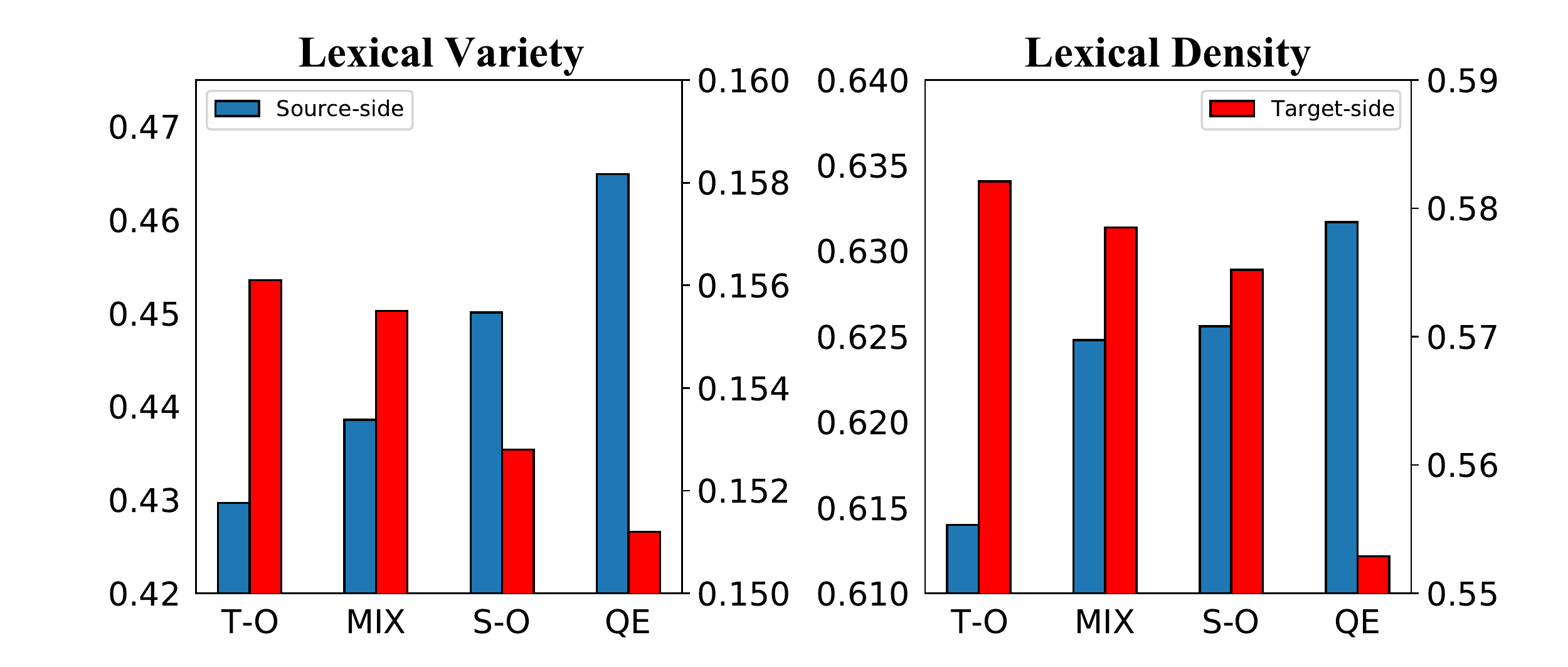}
\caption{\textbf{Comparison of stylistic features between target-original, mixed, source-original, and real QE bitext}, denoted by T-O, Mix, S-O, and Real QE respectively. The parallel sentences are randomly sampled from the WikiMatrix corpus and classified by our direction classifier in Section \ref{sec:classify}.}
\label{fig:stylistic}
\end{figure}

As Figure~\ref{fig:stylistic} displays, real QE data has the highest lexical variety and density on the source side (En) and lowest in the target side (Zh). \textit{Source-original bitext has closest lexical variety and density to QE data on both sides}, indicating that it has similar stylistic features with the gold-label dataset.


\paragraph{Domain Differences} As Figure~\ref{fig:motivation} shows, we can observe a gap between source- and target-original sentences inside the parallel corpus. Moreover, we also investigated content-dependent features from the view of vocabulary distributions. 
Specifically, we use the Jensen-Shannon (JS) divergence~\citep{Lin:1991:TIT} to measure the difference between target side vocabulary distributions of different bitext:
\begin{equation*}
 \mathrm{JS}\left (p || q \right ) = \frac{1}{2} (
\mathrm{KL} (p || \frac{p+q}{2} ) + \mathrm{KL} (q || \frac{p+q}{2})
), \nonumber
\end{equation*}
where $\mathrm{KL}(\cdot||\cdot)$ is the KL divergence~\citep{KL:1951:information} of two distributions. 

\begin{table}[t]
\centering
\begin{tabular}{lccc}
\toprule[0.5mm] 
\multirow{2}{*}{Data}  & \multicolumn{3}{c}{\textbf{JS Divergence}} \\
& S-O &  Mix & T-O  \\ \midrule
QE training data  &0.275 &0.292 & 0.305     \\  
\bottomrule[0.5mm]
\end{tabular}
\caption{\textbf{JS divergence of vocabulary distribution} between real QE data and different types of bitext.}
\label{tab:js}
\end{table}
Results in Table~\ref{tab:js} demonstrate that source-original bitext is closer to QE data, indicating their language coverages are more similar. This confirms our hypothesis that \textit{vocabulary distribution of source-original bitext is closer to real QE data}.



\section{Distinguish Source-Original \& Target-Original Bitext}
\label{sec:classify}
Unfortunately, the WMT parallel training corpus does not provide the original language information of a given sentence pair. Thus, a classifier has to be trained to identify translationese from the original text while training a classifier still needs annotated data. 

\begin{figure}[t]
\centering
\includegraphics[scale=0.43]{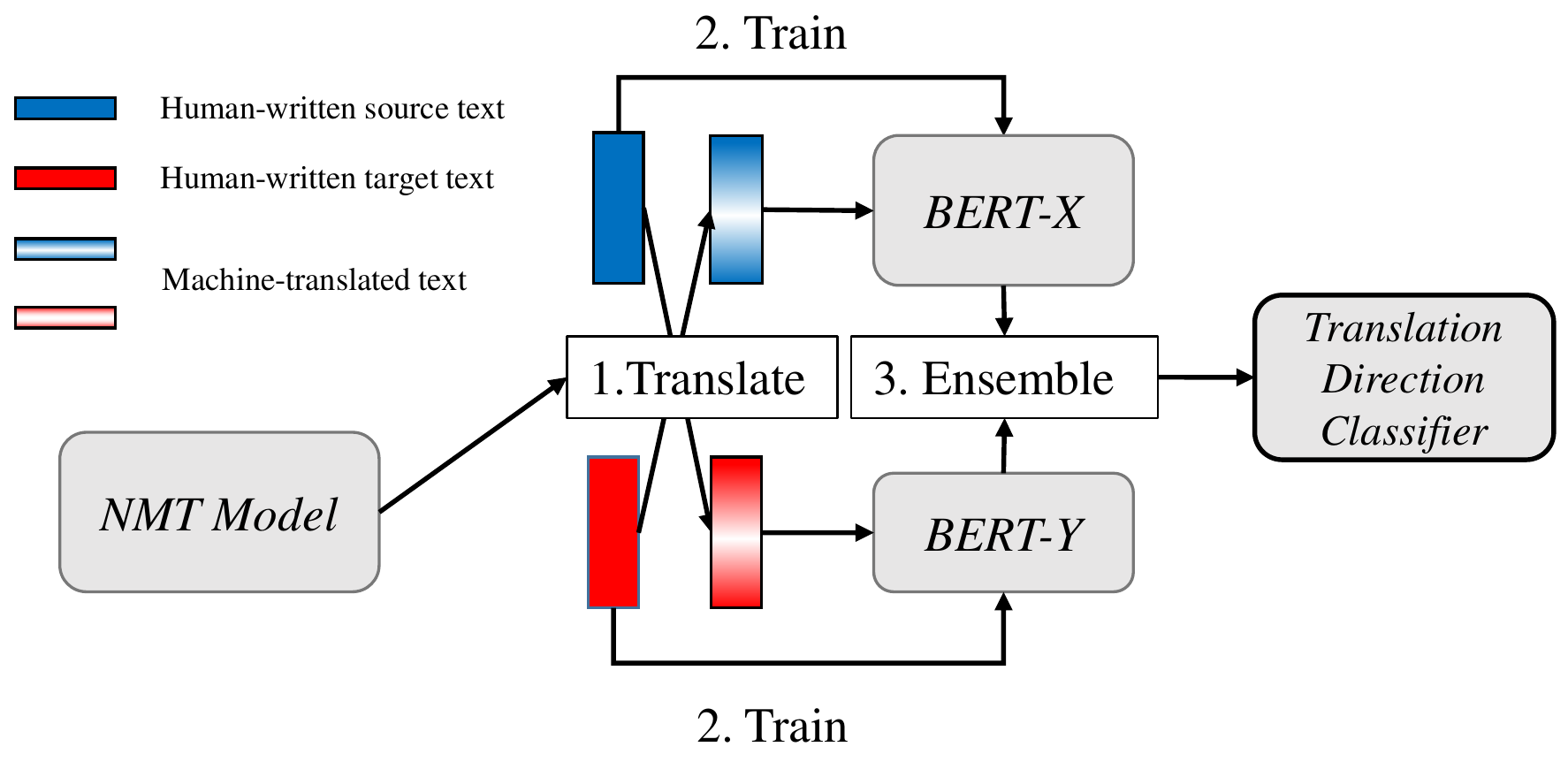}
\caption{\textbf{The translation direction classifier.} X and Y denote the source and target language respectively.}
\label{fig:classifier}
\end{figure}

As shown in Figure~\ref{fig:classifier}, our proposed model is an ensemble of two monolingual BERT-based classifiers \citep{DBLP:conf/naacl/DevlinCLT19} trained with fully synthetic data. We sample from WMT \emph{monolingual} corpus of source and target language respectively, and use an external NMT model to forward-translate the sampled sentences to \emph{translationese}. A target language BERT model is finetuned to classify translationese and original text. The process can also be performed accordingly for a source language BERT model.
We will explain details on our choice of model and training data below.


\subsection{Translationese Identifier: Training data}

Previous works \citep{riley2020translationese} prove that it is possible to train a translationese identifier using only monolingual data from source and target languages, since an MT system can be used as the proxy for humans to produce translated text (translationese). Translated text and original monolingual data in the target language can be used as negative/positive examples to train a classifier. 

To further improve classification performance, we investigate the following research questions: 

\noindent\textbf{Q1}. Does synthetic text with more translationese features serve as better negative samples that help in training the classifier? 

\noindent\textbf{Q2}. Is distinguishing source-/target-original bitext relied more on the identification of stylistic features, or content-dependent features? 

\begin{table}[t]
\small
\centering
\resizebox{\linewidth}{!}{
\begin{tabular}{llllc}
\toprule[0.5mm] 
\multicolumn{2}{l}{Method}  & En-De      &  En-Ja     & En-Zh     \\ \midrule
  \multicolumn{2}{l}{\citet{riley2020translationese} (82M) }             &   86.6    &   83.7    &    83.6  \\
  \multicolumn{2}{l}{\citet{DBLP:conf/acl/WangTTSSL21} (82M)}    &   88.7    &    91.5   & 84.4     \\ \midrule
  
 \multirow{3}{*}{BERT-tgt+} 
            &  $win=3$ (2M)                  &  79.7     & 82.1      &  77.4    \\
            &  RTT (2M)                 &  -     &  -     &  50.6    \\
            &  FT (2M)                 &   88.0    & 90.9      &  86.6    \\ \midrule
            
 \multirow{3}{*}{FT (2M)+}   &BERT-en    &    84.2   & 87.5      &   83.3   \\ 
     & XLM-R    & -      &  -     &   51.8   \\ 
 & BERT-ensemble   &  \textbf{89.2}     &  \textbf{91.8}     &  \textbf{87.9}    \\ 
\bottomrule[0.5mm]
\end{tabular}}
\caption{\textbf{Experimental results on translation direction classification}, evaluated on WMT20 test sets, measured by macro F1(\%). Settings are elaborated in Appendix \ref{sec:appendix}.}
\label{tab:model}
\end{table}

For \textbf{Q1},  \citet{DBLP:conf/iwslt/BizzoniJECGT20} find that in speech translation, human interpreters present more evident translationese symptoms due to cognitive limits, i.e. they tend to produce the word-for-word translation. Thus, we let our MT system mimic this process by following  \citet{DBLP:conf/coling/DingWWTT20} to constrict the attention scope in the self-attention module of all encoder layers, which can be formulated as:
\begin{equation*}
\label{clauserepresentation}
C(\psi_{i,j})=\left\{
\begin{aligned}
\psi_{i,j} &,  & i-win\leq j\leq i+win, \\
-\infty &, & otherwise.
\end{aligned}
\right.
\end{equation*}

\noindent where $\psi_{i,j}$ is the attention correlation between i-th and j-th element, which are aligned elements with the highest attention weight. To balance translation quality and severity of the translationese effect, we set $win=3$ heuristically.

The constrained model tends to decode the source text word-for-word, with more interference from the syntactic structure of the source language, and thus produce more evident translationese. As Table \ref{tab:encoder3} shows, machine-generated translation with constrained attention has less lexical variety and density than normal MT output, indicating a higher degree of translationese. However, as shown in Table~\ref{tab:model}, training a classifier with the designed translationese hurts model performance. Therefore, the answer to \textbf{Q1} is "No" based on our findings. 



\begin{table}[t]
\small
\centering
\resizebox{\linewidth}{!}{
\begin{tabular}{lcccc}
\toprule[0.5mm] 
\textbf{Data Type} & $win=3$ &  FT & RTT & \makecell[c]{Original\\Text}  \\ \midrule
Lex. Diversity  &0.0629 &0.1097 & 0.1128  & 0.236   \\  
Lex. Density  &0.5384 & 0.5507 & 0.5521  & 0.5517   \\  
\bottomrule[0.5mm]
\end{tabular}}
\caption{\textbf{Lexical variety \& density of translated text} under different settings.}
\label{tab:encoder3}
\vspace{-5pt}
\end{table}

To further explore the reasons, we raise \textbf{Q2}, and investigate it by constructing synthetic training data using round-trip translation (RTT). In RTT, target monolingual data $Y$ is translated to the source language, and then back-translated to produce $Y^{**}$. Compared with FT, $Y^{**}$ produced by RTT has exactly the \emph{same} content with $Y$, while the only difference lies in stylistic features, i.e. translationese. Unfortunately, results in Table~\ref{tab:model} show that BERT classifier trained with RTT data obtains extremely bad results, confirming the answer of \textbf{Q1}, while giving \textbf{Q2} also a negative answer that content-dependent features play a more important role in distinguishing source- and target-original bitext.

  
            

\subsection{Translationese Identifier: Model Choice}
As for model choice, there are works that utilize both sides of bitext \citep{DBLP:conf/acl/WangTTSSL21} by training two language models. However, training from scratch is costly in both data and time. Based on our findings, we focus on content-dependent features and leverage LMs pretrained with a large amount of monolingual data in a specific language. As Table~\ref{tab:model} shows, BERT-tgt (target language monolingual BERT) finetuned with only 2M synthetic FT data outperforms previous works that used 82M data, and the performance can be further boosted by ensembling BERT-tgt with BERT-en. We also tried to integrate both sides into a single cross-lingual PLM, XLM-R, but obtained poor results. We assume that it is because XLM-R is pretrained with multilingual data of various languages, it cannot utilize language-specific content features, leading to unsatisfying results. 

\section{Improving QE with Source-original}




We perform our experiments on the En-Zh dataset of WMT20 QE subtask 2 and evaluate the results on validation sets using the official evaluation scripts \citep{DBLP:conf/wmt/SpeciaBFFCGM20}. Results on other language pairs and specific datasets are detailed in Appendix~\ref{sec:res}.

\begin{figure}[ht]
\centering
\includegraphics[scale=0.33]{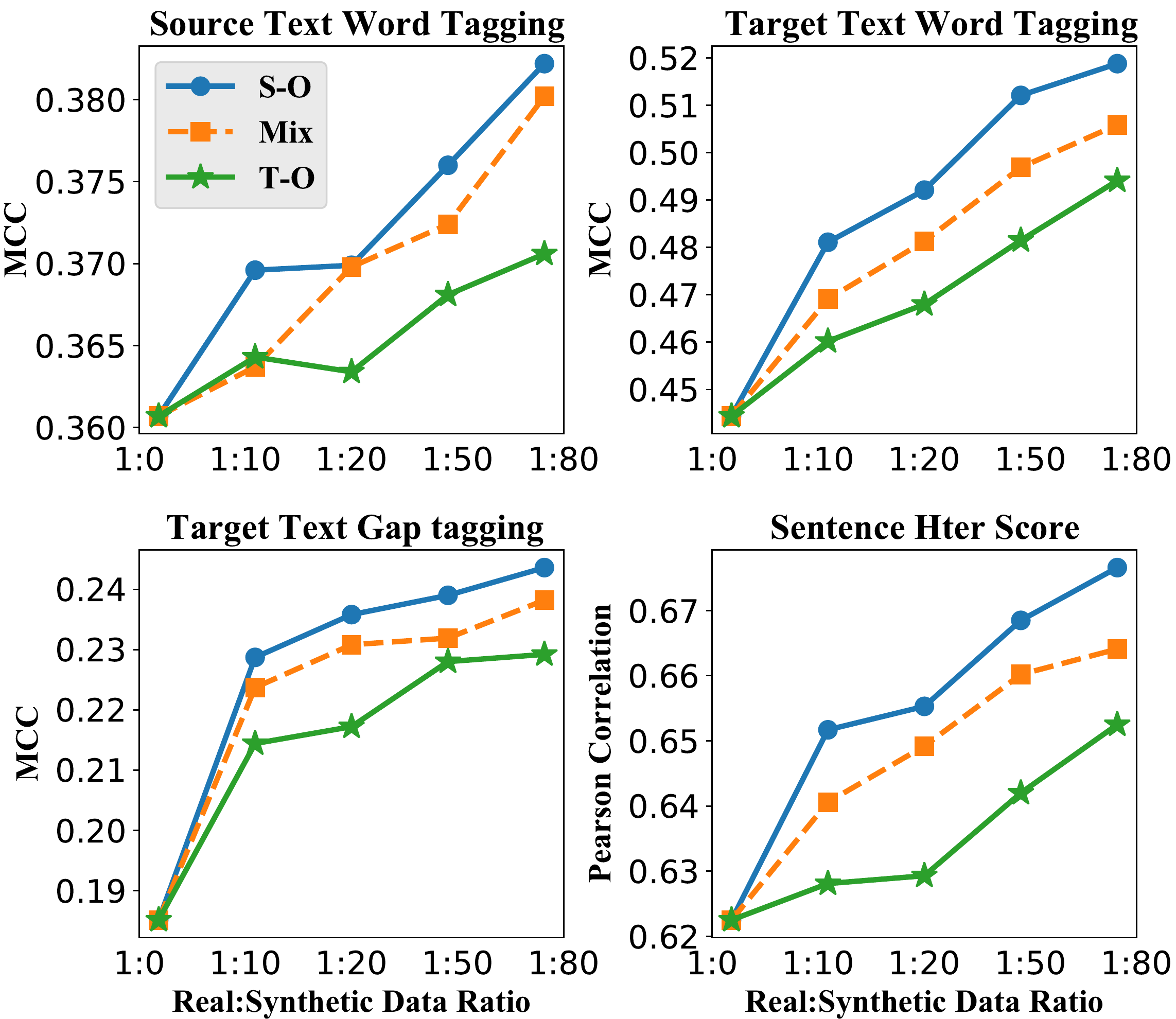}
\caption{\textbf{Results of QE data augmentation using different bitext}, evaluated on sentence-level task measured by Pearson correlation and word-level tagging task measured by Matthews correlation coefficient (MCC). Data ratio denotes the ratio of real and synthetic QE data.}
\label{fig:main}
\end{figure}

We find that QE models augmented with source-original bitext consistently outperform target-original or mixed bitext under different data ratio settings, especially for sentence-level HTER score prediction and word-level target-side word tagging with a max relative improvement of 4\% and 6.4\% respectively. The results on source text word tagging seem to be closer, since the standard QE data augmentation \citep{lee-2020-two} pipeline does not generate direct pseudo labels for source words, and this may be improved in future works.

Source-original bitext also presents data efficiency for QE data augmentation. We can observe from Figure~\ref{fig:main} that in the sentence-level task, source-original bitext, with a data ratio of 1:10, achieves similar performance with only 20\% amount of target-original bitext or 50\% amount of randomly selected bitext.

\section{Conclusion}
In this paper, we investigate the effect of different types of parallel data used in QE data augmentation. We propose a bitext classifier based on monolingual PLMs leveraging content-dependent differences between source and target languages. After distinguishing the parallel corpus, we conduct experiments on different QE tasks, and find that leveraging source-original bitext, which is naturally closer to real QE data, helps in most cases.

\section*{Limitations}
Previous works on translationese have claimed that sentence pairs in the parallel corpus are either source- or target-original, and use a binary classifier to distinguish translation directions, which is followed by our paper. However, statistics in Figure~\ref{fig:para} show there are some undirectional parallel data. And even in some highly-directional datasets, there are some sentences that may be originally written in a specific language but do not carry language-specific features. Future work may consider deciding whether the sentence pair is directional at first and then distinguishing their translation direction.

\bibliography{arxiv_version}
\bibliographystyle{acl_natbib}

\appendix

\section{Appendix}
\label{sec:appendix}
\subsection{Experimental Settings}
For training the classifier, we follow \citet{riley2020translationese} to use News Crawl for monolingual data (2017 for En, 2019 for Zh). For FT and RTT, we employ off-the-shelf translation models by importing \textsf{MarianMT} module from \textsf{transformers}. For the classifier, we use \textsf{bert-base-chinese} for Zh, \textsf{bert-base-uncased} for En \textsf{bert-base-japanese-whole-word-masking} for Ja, \textsf{bert-base-german-cased} for De, and train the model for 2 epochs. 

For QE data augmentation, we distinguish source- and target-original bitext in the WMT22 En-Zh parallel corpus. To balance over parallel datasets of different domains, we randomly sample \emph{an equal number of} sentence pairs from each dataset, and produce synthetic data by computing TER for sentence-level labels and word-level tags. The QE model is based on \textsf{xlm-roberta-large}. At the QE pretraining stage, we combine all sentence-level and word-level loss and  train for 2 epochs, then finetune with gold label QE training data of specific subtasks for 10 epochs. 




\subsection{Detailed Results}
\label{sec:res}
\subsubsection{Generalization of Bitext Classifier}
\begin{table}[ht]
\small
\centering
\begin{tabular}{lccc}
\toprule[0.5mm] 
Dataset & BERT-zh & BERT-en & Ensemble  \\ \midrule
WMT17  &88.1 &86.8 &90.8       \\  
WMT18    & \textbf{85.3}  & 79.5  & \textbf{85.3}     \\ 
WMT19    &85.5   &85.9    &\textbf{89.4}  \\ 
WMT20  &86.6   &83.3   &\textbf{87.9}  \\ 
WMT21  &67.7    &71.1     &\textbf{73.9}  \\ 
\bottomrule[0.5mm]
\end{tabular}
\caption{Classification results of src- and tgt-original text in more test sets, measured by macro F1(\%).}
\label{tab:mainresults}
\end{table}

\subsubsection{Results on En-De}
\begin{table}[ht]
\small
\centering
\begin{tabular}{lccc}
\toprule[0.5mm] 
\multirow{2}{*}{QE task}  & \multicolumn{3}{c}{\textbf{Bitext Type}} \\
& S-O &  Mix & T-O  \\ \midrule
Sentence-level HTER  &0.588 &0.609 & 0.628     \\  
Target word tagging  &0.464 &0.517 & 0.531     \\  
Source word tagging  &0.361 &0.367 & 0.385     \\  
\bottomrule[0.5mm]
\end{tabular}
\caption{QE results on WMT20 En-De development set, augmented by parallel data randomly generated from WMT22 En-De parallel corpus.}
\label{tab:ende}
\end{table}
\subsubsection{Results on WikiMatrix}
\begin{table}[ht]
\small
\centering
\begin{tabular}{lccc}
\toprule[0.5mm] 
\multirow{2}{*}{QE task}  & \multicolumn{3}{c}{\textbf{Bitext Type}} \\
& S-O &  Mix & T-O  \\ \midrule
Sentence-level HTER  &0.676 &0.650 & 0.627     \\ 
Target word tagging  &0.512 &0.473 & 0.442     \\ 
Source word tagging  &0.358 &0.364 & 0.367     \\ 
\bottomrule[0.5mm]
\end{tabular}
\caption{QE results on WMT20 En-Zh development set, augmented by parallel data randomly generated from WikiMatrix corpus.}
\label{tab:wiki}
\end{table}

\end{document}